# Enhancing efficiency in paediatric brain tumour segmentation using a pathologically diverse single-centre clinical dataset

Arianna Piffer [1], Josef Alois Buchner [2,3,4], Antonio Giulio Gennari [5,6], Patrice Grehten [7,] Selma Sirin [7], Elizabeth Ross [8], Ivan Ezhov [9,10], Marcel Rosier [11,12], Jan Caspar Peeken [2], Marie Piraud [11], Bjoern Menze [12], Ana Guerreiro Stücklin [1], Andras Jakab [6,13*], Florian Kofler [10, 11, 12, 14*]

1 Division of Oncology and Children's Research Center, University Children's Hospital Zurich, 8008, Zurich, Switzerland; (AP.; A.G.S.)

2 Department of Radiation Oncology, TUM School of Medicine and Health, Klinikum rechts der Isar, Technical University of Munich, Munich, Germany (J.A.B.; J.C.P.)

3 Institute of Radiation Medicine (IRM), Helmholtz Zentrum, Oberschleißheim, Germany (J.A.B.)

4 Partner Site Munich, German Consortium for Translational Cancer Research (DKTK), Munich, Germany (J.A.B.)

5 Department of Neuropaediatrics, University Children's Hospital Zurich, 8008, Switzerland (A.G.G.)

6 Center for MR- Research, University Children's Hospital Zurich, Zurich, Switzerland (A.G.G., A.J.)

7 Department of Diagnostic Imaging, University Children's Hospital Zurich, Zurich 8008, Switzerland (P.G.; S.S.)

8 Georgia Institute of Technology; Geisel School of Medicine at Dartmouth (E.R)

9 Department of Computer Science, TUM School of Computation, Information and Technology, Technical University of Munich, Munich, Germany (I.E.)

10 TranslaTUM - Central Institute for Translational Cancer Research, Technical University of Munich, Munich (I.E., F.K.)

11 Helmholtz AI, Helmholtz Zentrum Munich, Munich, Germany (M.R., M.P., F.K.)

12 Department of Quantitative Biomedicine, University of Zurich, Zurich, Switzerland (M.R., B.M., F.K.)

13 Faculty of Medicine, University of Zürich, Switzerland (A.J.)

14 Department of Diagnostic and Interventional Neuroradiology, Klinikum rechts der Isar, Technical University of Munich, Munich, Germany (F.K.)

Corresponding author: Arianna Piffer, Lenngstrasse 30, +41788503443, Arianna.piffer@kispi.uzh.ch

**shared last authorship*

**Total Word Count**

- Abstract: 245
- Main Text: 4792
- References: 1173
- Figure and Table Legends (420):
    - Figure 1: 53
    - Table 1: 16
    - Figure 2: 70
    - Table 2: 32
    - Figure 3: 102
    - Figure 4: 53

## Abstract

**Background** Brain tumours are the most common solid malignancies in children, encompassing diverse histological, molecular subtypes and imaging features and outcomes. Paediatric brain tumours (PBTs), including high- and low-grade gliomas (HGG, LGG), medulloblastomas (MB), ependymomas, and rarer forms, pose diagnostic and therapeutic challenges. Deep learning (DL)-based segmentation offers promising tools for tumour delineation, yet its performance across heterogeneous PBT subtypes and MRI protocols remains uncertain.

**Methods** A retrospective single-centre cohort of 174 paediatric patients with HGG, LGG, medulloblastomas (MB), ependymomas, and other rarer subtypes was used. MRI sequences included T1, T1 post-contrast (T1-C), T2, and FLAIR. Manual annotations were provided for four tumour subregions: whole tumour (WT), T2-hyperintensity (T2H), enhancing tumour (ET), and cystic component (CC). A 3D nnU-Net model was trained and tested (121/53 split), with segmentation performance assessed using the Dice similarity coefficient (DSC) and compared against intra- and inter-rater variability.

**Results** The model achieved robust performance for WT and T2H (mean DSC: 0.85), comparable to human annotator variability (mean DSC: 0.86). ET segmentation was moderately accurate (mean DSC: 0.75), while CC performance was poor. Segmentation accuracy varied by tumour type, MRI sequence combination, and location. Notably, T1, T1-C, and T2 alone produced results nearly equivalent to the full protocol.

**Conclusions** DL is feasible for PBTs, particularly for T2H and WT. Challenges remain for ET and CC segmentation, highlighting the need for further refinement. These findings support the potential for protocol simplification and automation to enhance volumetric assessment and streamline paediatric neuro-oncology workflows.

## Key Points

- The public released model achieves high accuracy in WT and T2H, moderate on ET and limited on CC
- Automatic segmentations reduced manual contouring time by up to 83%
- Optimized MRI protocols preserve performance, enabling shorter scans

## Importance of the Study

PBTs require precise segmentation for treatment planning and response assessment. Manual segmentation is labour-intensive and variable, necessitating automated solutions. While DL models have shown success in adult tumours, their performance on paediatric cases is limited due to anatomical differences, fewer cases and tumour heterogeneity. This study presents a DL-based segmentation model trained on single-centric, pathologically diverse paediatric MRI data, demonstrating high accuracy for WT and T2-hyperintensity segmentation. By quantifying time savings and evaluating MRI sequence contributions, our work provides not only a segmentation tool, but also practical guidance for optimizing imaging workflows in the paediatric setting.

## Main Text

## *1. Introduction*

PBTs are the most common solid cancers in children and the leading cause of cancer-related child mortality(1). While they may share certain features with adult brain tumours, PBTs show distinct biological and radiological characteristics, including variations in histologic subregions such as the enhancing tumour core and cystic component. Accurate characterization and segmentation of these subregions are important for treatment planning, response assessment, and for assessing prognosis[2].

With the advent of novel therapies for PBTs (e.g. target therapies), precise and reproducible methods for evaluating treatment response or relapse are essential. Traditional response assessment methods rely on 2D measurements, which assume uniform tumour growth(3); however, this may not accurately capture the complex, often irregular growth patterns of PBTs, moreover after therapy. Volumetric assessment through tumour segmentation provides a better estimate of the 3D volume and morphology of the growing tumour (4),(5),(6). This is important for PBTs, as they are generally irregularly shaped and often have complex, mixed solid and cystic components(3). The quantification of these subregions often requires manual segmentation, which is labour-intensive, time-consuming, and subject to inter-rater variability(7).

DL-based automatic segmentation has significantly advanced adult brain tumour analysis as reflected by the impact the Brain Tumour Segmentation (BraTS) Challenge made on the field, also leading to open-source software, such as BraTS toolkit(8–13),(14). However, models trained on adult tumours do not generalize well to PBTs due to anatomical differences in the developing paediatric brain and distinct tumour locations and radiological features(15,16). In adults, peritumoral oedema is frequently present and routinely segmented as a distinct region—especially in the context of standardized frameworks such as BraTS. However, in the paediatric setting, peritumoral oedema is less common in PBTs but can be present depending on histology.. Instead, in paediatric, T2-hyperintensity emerges as the most reliable marker for delineating tumour boundaries. T2H often corresponds to the full tumour extent and can be present even in the absence of contrast enhancement, which is common in several paediatric tumour subtypes. This observation underscores the need for adapted annotation protocols and segmentation strategies tailored to the paediatric population, where traditional adult-centric labels may be insufficient or misleading.

While some studies have proposed automated segmentation methods for PBTs, these models often demonstrate lower performance, are limited to specific histologies, or rely on a minimal number of MRI sequences or segment the whole tumour without considering subregions, restricting their clinical applicability(17–23). Previous efforts, such as BraTS-Peds and CHOP, have already demonstrated the feasibility of automated segmentation in PBTs, but these were constrained by narrower histological representation. We conclude that there remains a clear need for a comprehensive segmentation approach tailored to PBTs: one that is accurate, efficient, and generalizable across histological types and tumour locations, capable of delineating subregions, and integrating data from multiple MRI sequences. To address these challenges and to improve PBT quantification in the clinical practice, we propose a DL-based segmentation framework tailored specifically for PBTs by training it on a clinically and histologically diverse dataset. We evaluated segmentation accuracy across ages, tumour localizations, and histologies by analysing specific tumour subregions, comparing performance and reproducibility with human annotators, and assessing the contribution of each MRI sequence to accuracy. The latter step was done by evaluating models trained with different MRI sequence combinations to determine the optimal balance between accuracy and clinical feasibility(24).

## *2. Methods*

### *2.1 Patient cohort*

A retrospective collection of MRI scans was conducted, including only pre-treatment scans obtained at initial diagnosis from paediatric patients aged 0–18 years treated at the University Children's Hospital Zurich between 2014 and 2022. Patients with first available scan performed post-biopsy or external ventricular drain (EVD) placement without significant malformation of brain skull and/or parenchyma were also considered. Patients were excluded from the study if any of the four standard scans (native T1-weighted sequence (T1), post-contrast T1-weighted sequence (T1-C), T2-weighted sequence (T2), and T2 Fluid Attenuated Inversion Recovery (FLAIR)) were missing, or if the scans were not acquired prior to surgical resection. A total of 174 subjects met the inclusion criteria for this study (Figure 1, Table 1). 66% of patients presented an enhancing tumour with variable intensity (N = 109). 30% of patients presented a cystic component (N = 52), among which 40% showed an enhancing cyst wall.

Public datasets such as BraTS-Peds were not integrated, as they primarily include high-grade gliomas and diffuse midline gliomas, which would distort the histological distribution of our cohort. Moreover, BraTS-Peds images are skull-stripped, whereas we intentionally included non–skull-stripped scans to preserve extra-axial tumour regions—such as optic pathway gliomas—that are clinically relevant but located outside the brain parenchyma.

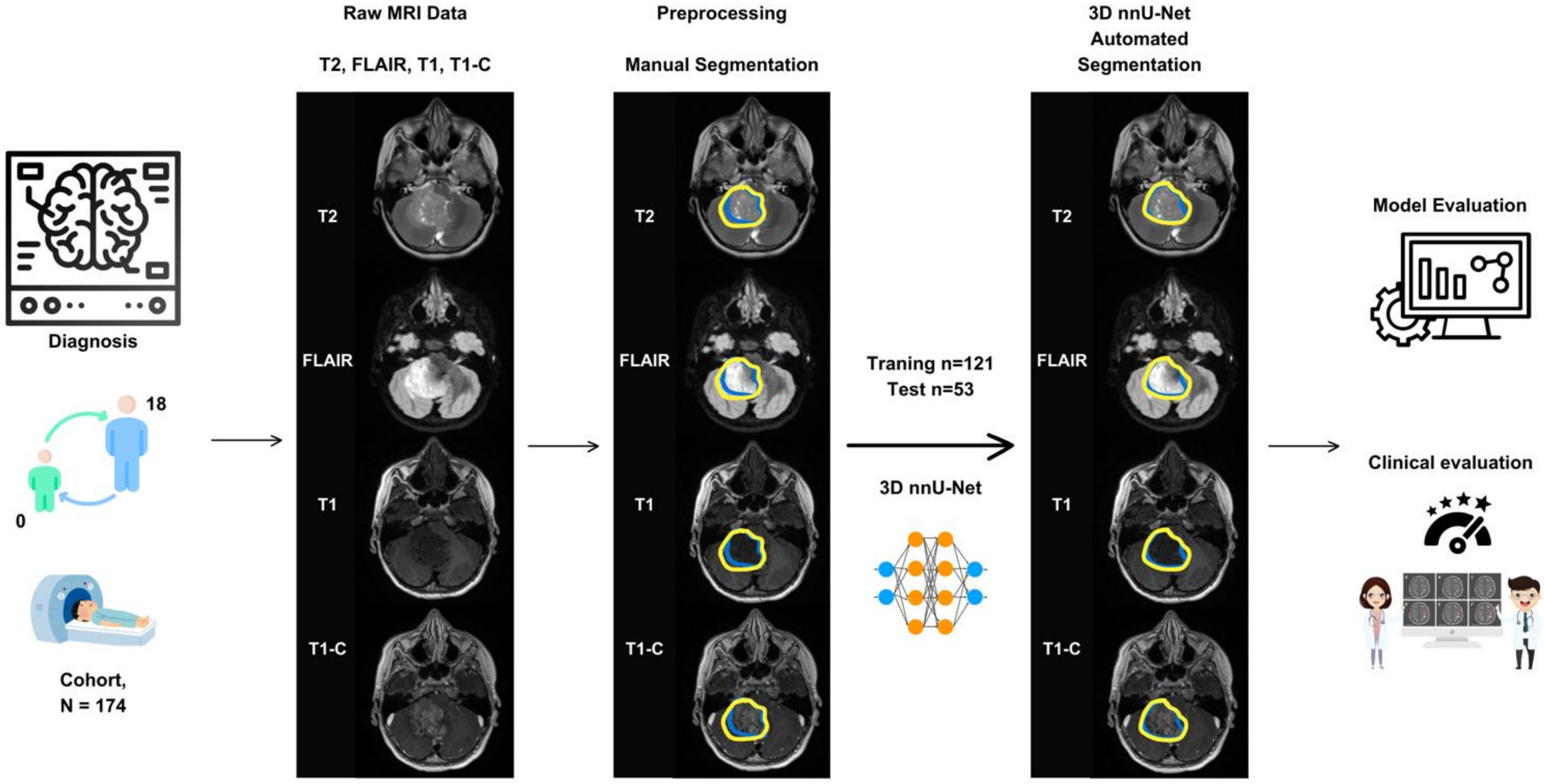

*Figure 1: Study workflow, including data collection, preprocessing, expert segmentation of the tumours, model training and testing and model evaluation. Statistical model evaluation includes the Dice score (DSC) as well as volume difference. The model clinical evaluation was based on 4 star rating (rejection, usable with major modifications, usable with minimal modifications, usable).*

| | Patients Cohort | |
|---|---|---|
| *Parameter* | *Training (n=121)* | *Test (n=53)* |
| **Age (y)** | | |
| Median (range) | 6 (2.75-10) | 6 (2.5 -10) |
| **Sex** | | |
| Female | 51 | 23 |
| Male | 70 | 30 |
| **Tumour Types** | | |
| Medulloblastoma | 20 | 11 |
| Ependymoma | 13 | 6 |
| High Grade/Diffuse Midline Glioma | 14 | 10 |
| Low Grade Glioma | 54 | 18 |
| Other | 20 | 8 |
| *Germ cell tumours* | *7* | *2* |
| *Craniopharyngioma* | *3* | *1* |
| *Meningioma* | *1* | *1* |
| *Pinealoblastoma* | *1* | *0* |
| *Papillary tumour of the pineal region* | *1* | *0* |
| *Choroid plexus papilloma* | *1* | *1* |
| *Hemangioblastom* | *1* | *1* |
| *Atypical Teratoid/Rhabdoid Tumours (AT/RT)* | *1* | *1* |
| *Not specific* | *4* | *1* |
| **Primary Tumour Location** | | |
| Brain Hemispheres | 21 | 8 |
| Posterior Fossa | 58 | 28 |
| Brainstem | 14 | 7 |
| Pinealis | 8 | 3 |
| Other | 20 | 7 |
| *Optic* | *5* | *2* |
| *Thalamus* | *7* | *3* |
| *Chiasma* | *4* | *1* |
| *Suprasellar* | *4* | *1* |

***Table 1.*** *Overview of patient demographics, tumour histology, and primary tumour locations in the study cohort*

### *2.2 Data pre-processing and tumour subregion segmentation*

For all patients, a standardized pre-processing approach was applied to MRI scans. First, files were converted from Digital Imaging and Communications in Medicine (DICOM) format to Neuroimaging Informatics Technology Initiative (NIfTI, gzipped) format, during which all patient identifiers were removed (Figure 1).

Second, a multistep co-registration was conducted. All MRI modalities were registered to their respective T1-C image, followed by a registration to the 1mm³ isotropic SRI-24 atlas space(25). All co-registrations were conducted as rigid registrations. The pre-processing pipeline is publicly available on (https://github.com/BrainLesion/preprocessing) (26) as part of the BrainLesion suite (https://github.com/orgs/BrainLesion/repositories) (27). Brain extraction by means of skull stripping was intentionally omitted to keep the optic pathway intact and to allow for tumour segmentation within this region.

Manual annotation was performed in 3D Slicer (Version 4.13.0) with three-dimensional axial, sagittal, and coronal views (28). A medical doctor with experience in paediatric neurooncology (AP) carried out the first annotations under the guidance and supervision of two board-certified neuroradiologists (AGG, PG). Next, the neuroradiologists additionally checked and manually adapted the final segmentations.

Due to the diverse histological subtypes and unique MRI characteristics of paediatric tumours, we modified the standard BraTS annotation protocol to better align with the MRI features of the most common tumour types. The annotated tumour subregions included T2H, ET, and CC. ET was characterized by areas with enhancement on T1-C compared to T1, CC typically exhibited a hyperintense signal (bright) on T2 and a hypointense signal (dark) on T1-weighted MRI.

### *2.3 Intra- and inter-rater variability*

To assess the variability in segmentation evaluation among human experts, we measured the intra- and inter-rater variability. We selected a total of 20 of the 174 patients based on tumour type and patient age from both the training and test sets for intra- and inter-rater variability analysis. Two radiologists independently segmented these cases, differentiating between the different subregions. Intra-rater variability was assessed by having each radiologist segment the same cases at different time points, while inter-rater variability was determined by comparing segmentations between the two radiologists. To quantitatively evaluate inter- and intra-expert variability between the two annotators we computed the DSC, and compared it against the model performance to determine whether the model was within the range of human inter- and intra-expert variability.

### *2.4 Neural network training*

We selected 53 of the 174 patients based on tumour type and patient age to create balanced training and test sets. Patients with lower-quality images (due to poor resolution, motion artefacts, braces, etc.), which resulted in segmentation maps with higher ambiguity, were included exclusively in the training set. We used nnU-Net version 2.5.2 to train our deep-learning segmentation models(29). After extracting the dataset fingerprint, a convolutional neural network (CNN) with skip connections is trained according to the selected default experiment planner in the high-resolution 3D configuration. This results in 1000 epoch training runs with a linear learning rate decay.

Our experiments were conducted on a workstation with an Intel 9940X CPU and two NVIDIA RTX 8000 GPUs using CUDA version 12.7. This resulted in training times of approximately 67s per epoch using a single GPU, accumulating to a total of ~20h for one training run.

The deep learning models are publicly available(30) as part of the BrainLesion suite (27).

### *Label postprocessing*

To eliminate isolated false positives, particularly in the ET and CC channels, we applied a size-based filtering postprocessing step. Therefore, we applied a connected component analysis (CCA) for all three channels using cc3d (31) with 26-connectivity. This defines voxels that touch each other with at least one corner as connected.

Subsequently, we removed connected components too small to influence clinical decision-making (32,33) using a threshold of 125 voxels corresponding to 0.125 mL. While we report metrics for the threshold of 125 voxels, the implemented postprocessing step is fully customizable in our public code repository, allowing users to define their own voxel threshold and connectivity rules depending on their application needs (e.g., research, clinical trials, surgical planning).

### *2.5 MRI sequences combination analyses*

To evaluate the potential contributions of each MRI sequence to the segmentation accuracy, we trained models for three categories of sequence combinations. First, we trained a model for each modality individually to quantify its predictive value. Second, we trained a model for all four sequences together. Last we trained models with different combinations of input sequences: T1-C only, T1 only, T2 only, FLAIR only, T1-C + T1 + T2, T1-C + T2 + FLAIR, T1-C + T1, T1-C + T2, T1 + T2, T1 + FLAIR, T1-C + FLAIR, T2 + FLAIR.

### *2.6 Performance analysis*

We evaluate model performance in two ways. First, we assess the accuracy of segmentations using volumetric metrics, such as the DSC. Second, we conduct an expert evaluation to assess the clinical utility of the segmentations.

### *Volumetric and accuracy assessment*

To compare the model predictions with the expert-curated reference segmentation we compute the DSC and volume difference. As for ET and CC, both false positive (network predicted a CC/ET when none was annotated) and false negative cases (network failed to predict a CC/ET present in the ground truth) were assigned a DSC of 0. We evaluate each label channel (T2H, ET, CC) individually and a combination of all, which we call WT inspired by BraTS(11). We compute similarity metrics with panoptica(34). To align with radiological nomenclature, we utilized the T2H label, as it provides an excellent evaluation of tumour volume and treatment response in paediatric patients.

### *Clinical acceptability testing*

To assess clinical acceptability (35) model predictions were further evaluated by three expert assigning a score from 1 to 4 stars for each segmentation of the test set across different subregions. For each of the 53 cases of the test set, each expert was presented with the three 3D nn-UNet segmentations: T2H, ET, CC. Experts (AGG, SS) were given instructions and asked to rate each of the three segmentations. The star ratings were defined as follows:

- **1 star**: The segmentation is completely incorrect/not in the right location.
- **2 stars**: The segmentation is in the correct location but requires significant modifications.
- **3 stars**: The segmentation is in the correct location but needs minor adjustments.
- **4 stars**: The segmentation is clinically usable and perfect.

### *Evaluation of contouring time*

The time required for segmentation was measured in a sample of 20 patients to estimate the performance improvement enabled by auto-contouring. Since, in clinical practice, AI-generated contours must always be reviewed and potentially modified by a neuroradiologist, the reduction in segmentation time achieved with auto-contouring was calculated as:

$$Relative\ reduction\ in\ time\ = \frac{\left(T(Cman) - T(CAI, adj)\right)}{T(Cman)}$$

where *T(Cman)* represents the time required for manual segmentation and *T(CAI,adj)* is the time needed to obtain an adequate segmentation after adjusting the AI-generated contours(36).

The analysis was conducted using the 3D Slicer software. Specifically, the segmentation time was recorded both without the assistance of auto-contouring and separately when modifying the segmentations generated by the neural network to achieve a satisfactory segmentation.

## *3. Results*

### *3.1 Whole tumour and subregion segmentation*

#### *3.1.1 Whole tumour segmentation*

The WT segmentation achieved a DSC represented as median (SD) **of 0.85 (0.21)** on the test set, indicating a strong agreement between the model's predictions and the ground truth segmentations. The **low volume difference** for WT segmentation (**median: 16.3%, SD: 23.5%**) suggests that the predicted volumes closely approximated the ground truth volumes. These results suggest that many patients required only limited manual adjustment of the whole tumour boundary for volume assessment. Performance metrics—DSC and Volume Difference - for the WT, T2H, enhancing tumour, and CC on the test set are summarized in the boxplot shown in Figure 2A.

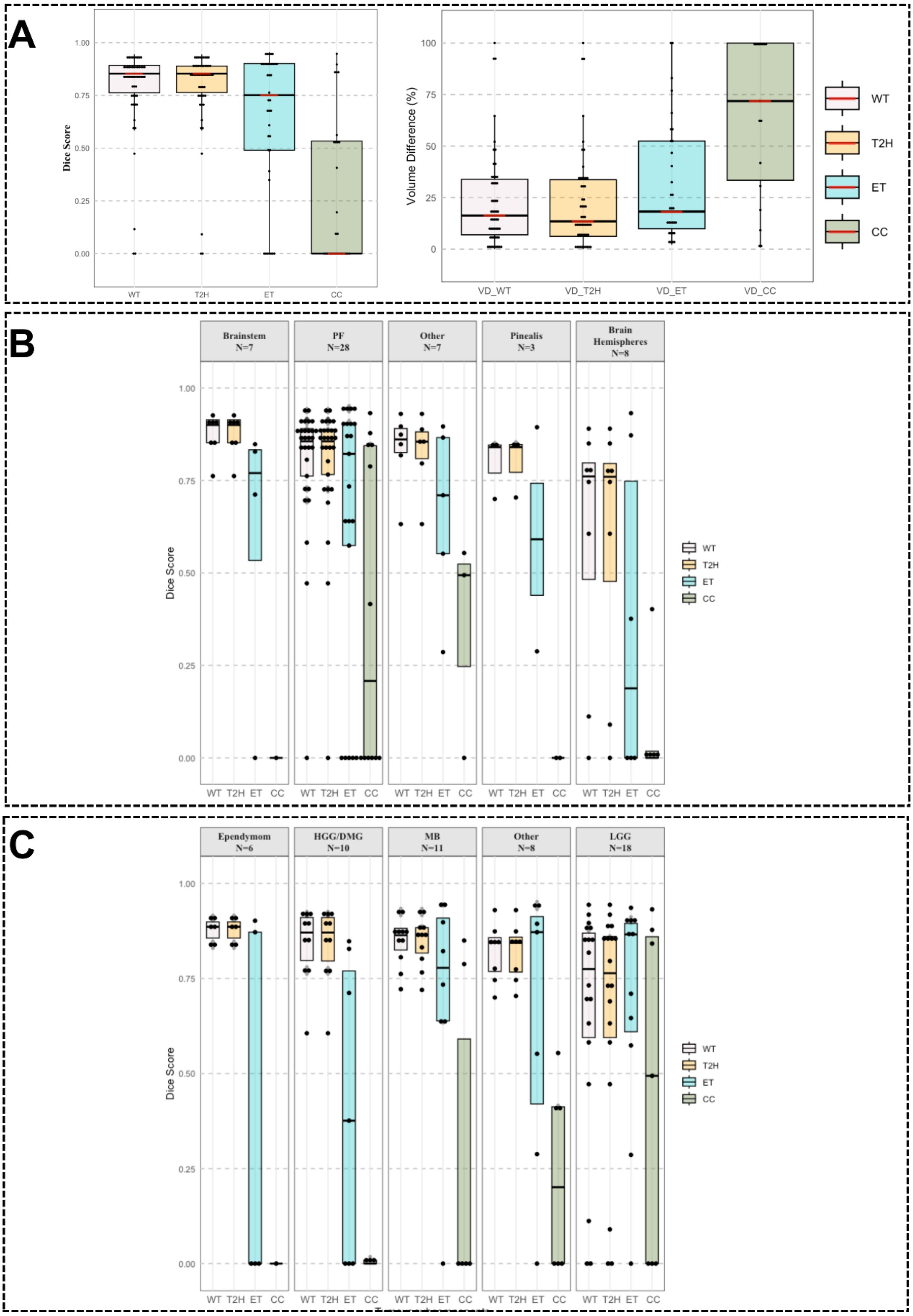

**Figure 2** Box plots summarizing the segmentation performance on the test set for T2H, ET and CC. (A) DSC distribution for different tumour subcomponents (left) and volume difference (right), averaged across all tumour types and locations. (B) DSC performance stratified by tumour localization. (C) DSC performance stratified by tumour type. *WT = whole tumour (combined mask); T2H = T2-hyperintense tumour region; ET = enhancing tumour component; CC = cystic component*

| | DSC (Median (SD)) | | |
|---|---|---|---|
| | *U Net* | *Intra-rater variability* | *Inter-rater variability* |
| **Tumour Types** | | | |
| Medulloblastoma | 0.86 (0.07) | 0.85 (0.12) | 0.83 (0.09) |
| Ependymoma | 0.89 (0.03) | 0.87 (0.06) | 0.87 (0.07) |
| High Grade/Diffuse Midline Glioma | 0.87 (0.1) | 0.88 (0.04) | 0.81 (0.08) |
| Low Grade Glioma | 0.82 (0.31) | 0.74 (0.2) | 0.72 (0.08) |
| Other | 0.85 (0.07) | 0.82 (0.08) | 0.78 (0.11) |
| **Primary Tumour Location** | | | |
| Brain Hemispheres | 0.76 (0.34) | 0.88 (0.1) | 0.86 (0.07) |
| Posterior Fossa | 0.86 (0.19) | 0.85 (0.2) | 0.83 (0.09) |
| Brainstem | 0.91 (0.06) | 0.89 (0.09) | 0.87 (0.09) |
| Pinealis | 0.84 (0.08) | 0.82 | 0.81 |
| Other | 0.88 (0.1) | 0.78 (0.03) | 0.78 (0.09) |

***Table 2.*** *DSC values (median and standard deviation) of WT (Whole Tumour) segmentation of the 3D U-Net compared to intra- and inter-rater variability, stratified by tumour histology and location*

### *3.1.2. T2 Hyperintensity segmentation*

To align with radiological nomenclature, we utilized the T2H as structure to segment. The model's DSCs for T2H closely matched those of WT segmentation, and the volume difference for T2H segmentation remained consistent with that of the WT (median: 13.5, SD: 23.5) (Figure 2A). This reinforces its suitability as an optimal metric for tumour volume calculation and response assessment. As WT is the combination of three separate subregion and the T2H region typically has the largest extent, the WT and T2H labels were often identical. Consequently, with a mean DSC of 0.85 and a SD of 0.21 on the test set, T2H segmentation demonstrated the same level of performance as WT segmentation (Table 2 and Figure 2).

| | DSC (Median (SD)) | | |
|---|---|---|---|
| | *U Net* | *Intra-rater Variability* | *Inter-rater Variability* |
| **Tumour Types** | | | |
| Medulloblastoma | 0.86 (0.07) | 0.84 (0.13) | 0.83 (0.09) |
| Ependymoma | 0.89 (0.03) | 0.88 (0.08) | 0.87 (0.07) |
| High Grade/Diffuse Midline Glioma | 0.87 (0.1) | 0.88 (0.08) | 0.81 (0.09) |
| Low Grade Glioma | 0.8 (0.31) | 0.74 (0.2) | 0.72 (0.08) |
| Other | 0.85 (0.07) | 0.82 (0.08) | 0.78 (0.11) |
| **Primary Tumour Location** | | | |
| Brain Hemispheres | 0.76 (0.35) | 0.88 (0.1) | 0.86 (0.07) |
| Posterior Fossa | 0.86 (0.19) | 0.85 (0.2) | 0.83 (0.09) |
| Brainstem | 0.91 (0.06) | 0.89 (0.09) | 0.87 (0.09) |
| Pinealis | 0.84 (0.08) | 0.82 | 0.81 |
| Other | 0.86 (0.1) | 0.78 (0.03) | 0.78 (0.09) |

***Table 3.*** *DSC values (median and standard deviation) of T2H (T2 Hyperintensity) segmentation of the 3D U-Net compared to intra- and inter-operator variability, stratified by tumour histology and location*

### *3.1.3. Enhancing tumour subregion segmentation*

The model demonstrated good performance in enhancing tumour segmentation, achieving a DSC of 0.75 (0.34) and a volume difference of 18% (SD = 29) on the test set (Figure 2A). However, the variability in performance across the enhancing component could be attributed to the differing levels of enhancement. For instance, the network performance was positively correlated with the intensity of contrast enhancement (Figure 3A), but underperformed with more mildly enhancing regions (Figure 3B), where its segmentation accuracy was lower. This variability is also seen in the inter-variability of our experts.

| | **DSC (Median (SD))** | | |
|---|---|---|---|
| | *U Net* | *Intra-rater Variability* | *Inter-rater Variability* |
| **Tumour Types** | | | |
| Medulloblastoma | 0.76 (0.32) | 0.82 (0.06) | 0.85 (0.4) |
| Ependymoma | 0 (0.5) | 0.66 (0.3) | 0.6 (0.43) |
| High Grade/Diffuse Midline Glioma | 0.49 (0.38) | 0.45 (0.4) | 0.65 (0.4) |
| Low Grade Glioma | 0.87 (0.3) | 0.84 (0.2) | 0.72 (0.38) |
| Other | 0.88 (0.19) | 0.7 (0.12) | 0.62 (0.39) |
| **Primary Tumour Location** | | | |
| Brain Hemispheres | 0.52 (0.34) | 0.68 (0.4) | 0.38 (0.4) |
| Posterior Fossa | 0.8 (0.39) | 0.76 (0.2) | 0.68 (0.2) |
| Brainstem | 0.8 (0.42) | 0.83 (0.04) | 0.75 (0.4) |
| Pinealis | 0.8 (0.43) | 0.68 | 0.78 |
| Other | 0.7 (0.3) | 0.74 | 0.6 |

***Table 4.*** *DSC values (median and standard deviation) of ET (Enhancing Tumour) segmentation of the 3D U-Net compared to intra- and inter-operator variability, stratified by tumour histology and location*

### *3.1.4. Cystic Component segmentation*

The CC demonstrated very low performance, with mean, median, and standard deviation DSCs of 0.26, 0, and 0.35, respectively. However, only 30% of cases contained a cystic component, which included two subtypes: ring-enhancing and non-enhancing cysts (Figure 3 C-D). The non-enhancing cysts posed greater segmentation challenges, contributing to the overall lower accuracy of the CC. This reduced performance can be attributed to the heterogeneity of the cystic component and its limited representation in the training data.
In the intra-variability the median and standard deviation of the DSCs were 0.8, 0.128 and in the inter-variability were 0.72, 0.224.

### *3.1.5. Segmentation performance across histologic subgroups*

Performance across histologic subgroups is reported in detail in Tables 2–4. Notably, segmentation accuracy was highest for ependymomas and lowest for LGG, underscoring the variability introduced by histologic heterogeneity. This analysis highlights the novelty of our work, as few previous studies have systematically compared performance across paediatric tumour histologies.

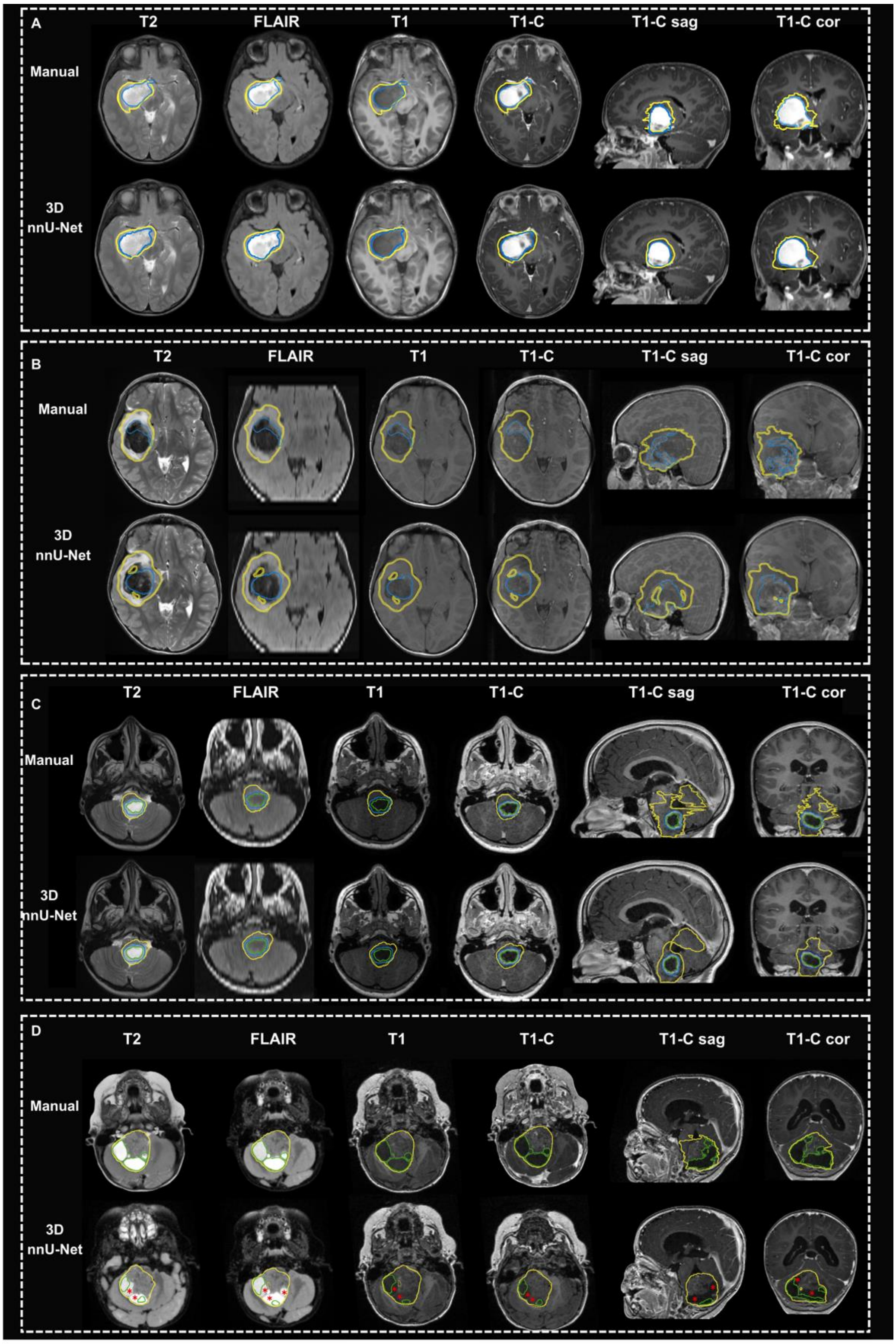

**Figure 3** Example tumours segmentations: cystic and enhancing component segmentation. Panel A shows enhancing components, with bright enhancement, manual annotations (top row) with 3DU-Net predictions (bottom row). Panel B shows enhancing components, with mild enhancement, manual annotations (top row) with 3DU-Net predictions (bottom row). The manual segmentations display a characteristic "Christmas tree effect" in the coronal and sagittal views — a visual artefact where isolated voxels appear stacked or scattered across slices, resembling a conical or tree-like pattern. *Outlines indicate different segmentation labels: yellow = T2H, blue = ET, green = CC. Red stars highlight false positive predictions generated by the 3D nnU-Net.*

### *3.1.6. Comparison with publicly available models*

To contextualize our results with respect to existing segmentation tools, we benchmarked our network against the first-ranked models from the BraTS Challenges 2023 and 2024. When applied directly to our skull-included, clinically diverse paediatric dataset, the BraTS-winning model achieved a WT Dice score of 0.11, confirming that current public models—trained on skull-stripped adult gliomas—fail to generalize to paediatric anatomy and heterogeneous tumour types. We used WT as the primary evaluation label for this comparison, as our annotation protocol differs from the BraTS definition of tumour subregions.

As a cross-validation, we ran the same BraTS-winning code on the BraTS training set and reproduced the expected WT performance (0.79). Conversely, when we applied our model—never trained on skull-stripped images—to the BraTS dataset, it achieved a WT Dice of 0.69, despite being tested only on HGG cases. This demonstrates that our network exhibits greater adaptability and robustness across anatomical and histological variations compared to current state-of-the-art BraTS models.

### *3.2 Evaluation of clinical utility of segmentations*

We implemented a Likert scale rating system to better reflect the clinical utility of our model's segmentations(35). Expert 1 assigned the following scores, presented as mean (SD): for T2H DSCs were 3.5 (0.67), for ET 3.14 (0.87), for CC 1.94 (1). Expert 2's scores were: for T2H DSCs were 3.58 (0.6), for ET 3.12 (1), for CC 2 (1.13) (Figure 4).

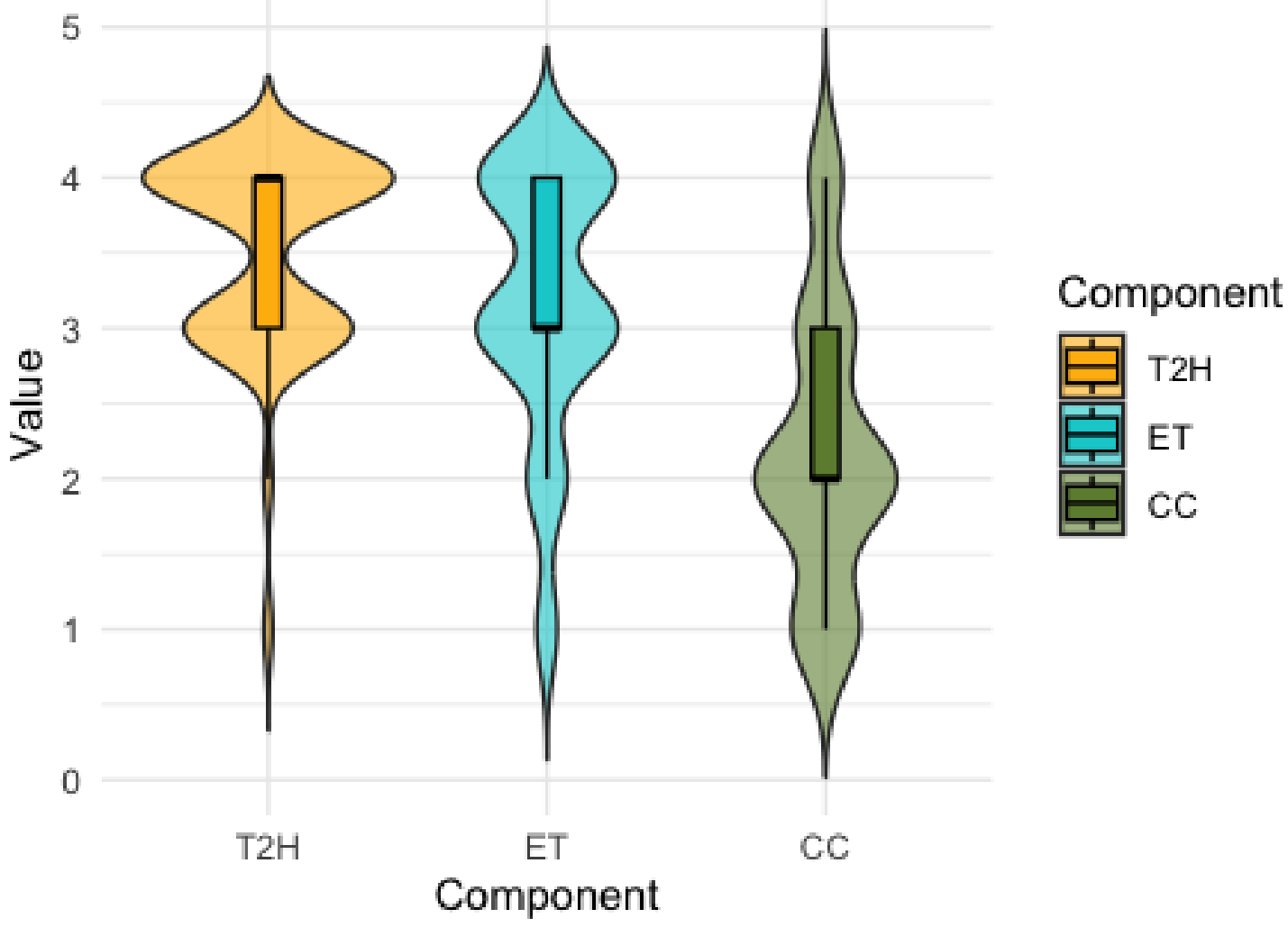

**Figure 4.** Clinical acceptability ratings of automated segmentations - Violin plot with overlaid boxplots showing the distribution of segmentation metrics for the individual tumour components T2H, ET, and CC. Ratings were provided on a 1–5 Likert scale (y-axis). *T2H = T2-hyperintense tumour region; ET = enhancing tumour component; CC = cystic component.*

### *Time saving in clinical routine*

To assess the clinical impact of automated segmentation, we compared manual and AI-assisted contouring times across all test subjects. For T2H, manual segmentation required a median of 48 minutes and 25 seconds, while AI-assisted segmentation reduced this to 8 minutes and 14 seconds—yielding a time saving of 40 minutes and 11 seconds (83%). For ET, the time was reduced from 26 minutes and 21 seconds to 15 minutes and 12 seconds, corresponding to a saving of 11 minutes and 9 seconds (42%). For CC, segmentation time decreased from 22 minutes and 54 seconds to 16 minutes and 48 seconds, saving 6 minutes and 6 seconds (27%). These results indicate substantial time reductions with AI assistance, particularly for T2H, and support the integration of automated segmentation tools to improve efficiency in clinical workflows.

| | DSC (Median (SD)) | | |
|---|---|---|---|
| | *Tman* | *T AI, adj* | *Time Savings* |
| T2H | 48 min 25 s | 8 min 14 s | 40 min 11 s (83%) |
| ET | 26 min 21 s | 15 min 12 s | 11 min 9 s (42%) |
| CC | 22 min 54 s | 16 min 48 s | 6 min 6 s (27%) |

**Table 5.** Average time required for manual (Tman) and AI-assisted (T AI, adj) contouring across all test subjects, with absolute and percentage time savings. *WT = whole tumour (combined mask); T2H = T2-hyperintense tumour region; ET = enhancing tumour component; CC = cystic component.*

### *3.3 Segmentation performance in various MRI sequence combinations*

Based on previous work, which investigated the optimal MRI sequence combinations for automatic segmentation, emphasizing the importance of reducing scan time by omitting redundant sequences without compromising segmentation performance(24), which could be particularly valuable for paediatric imaging.

We evaluated the DSCs for different tumour subregions and WT segmentation across various MRI sequence combinations (Figure 5, Table 5).

In summary, T1-C+ T1 + T2 showed the best performance for segmenting WT, T2H, and ET subregions, with reasonably high DSCs (WT: 0.84, T2H: 0.84, ET: 0.69).

Moreover, across sequence combinations, WT performance showed only minimal variation (median DSC range: 0.81–0.86). Differences between combinations were not statistically significant, suggesting that reduced sequence sets (e.g., T1, T1-C, and T2) may be sufficient for robust clinical application.

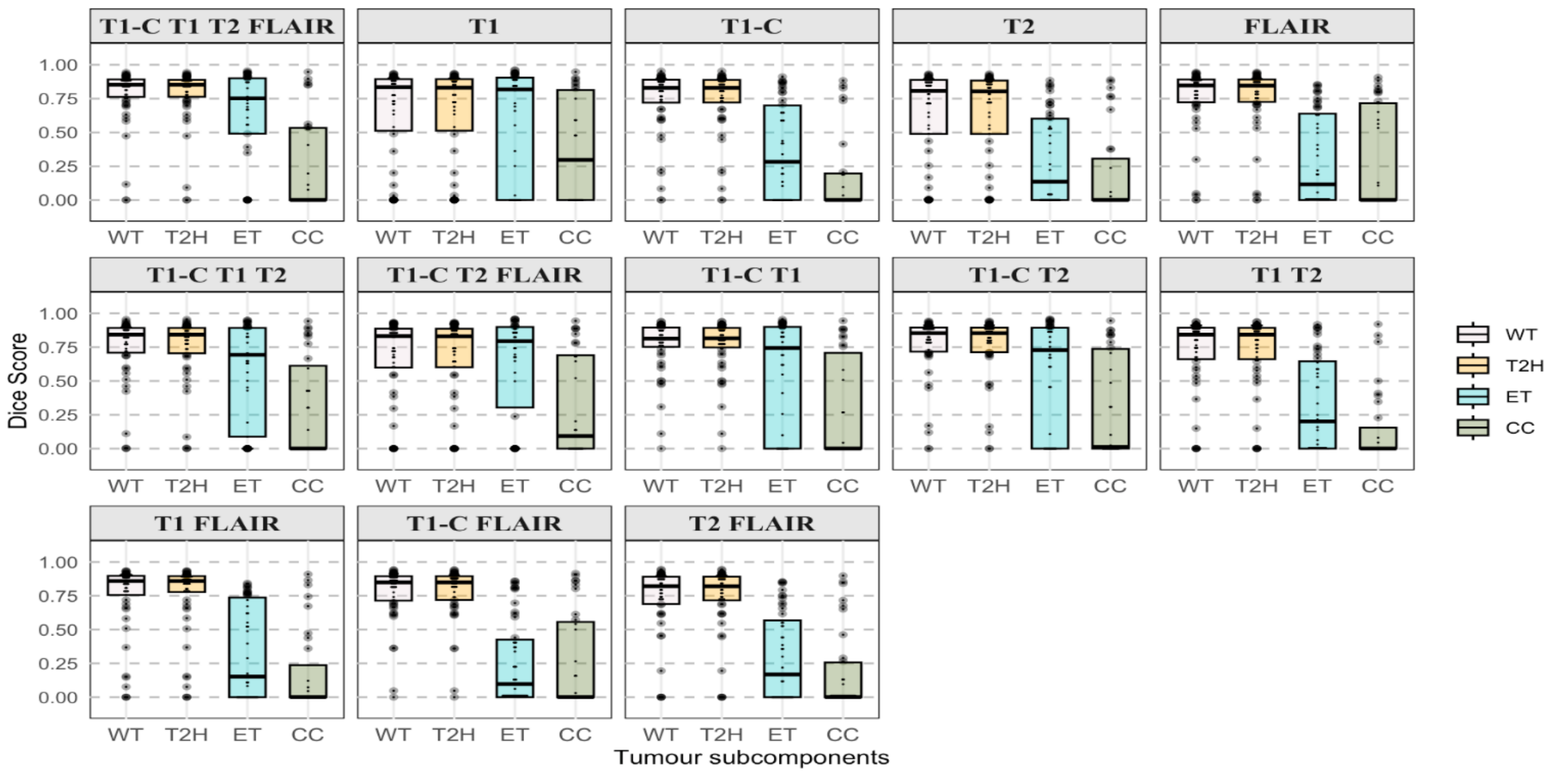

**Figure 5**. DSC performance on the test set for models trained with different combinations of MRI sequences. Each box plot represents the segmentation accuracy for different tumour subcomponents (T2-hyperintense, enhancing tumour, and cystic components) across various MRI input configurations. *WT = whole tumour (combined mask); T2H = T2-hyperintense tumour region; ET = enhancing tumour component; CC = cystic component.*

| | DSC (Median (SD)) | | |
|---|---|---|---|
| | *T2H* | *ET* | *CC* |
| T1-C + T1 + T2 + FLAIR | 0.85 (0.21) | 0.75 (0.34) | 0 (0.35) |
| T1-C | 0.83 (0.34) | 0.81 (0.4) | 0.29 (0.38) |
| T1 | 0.83 (0.24) | 0.28 (0.35) | 0 (0.35) |
| T2 | 0.81 (0.33) | 0.14 (0.33) | 0 (0.32) |
| FLAIR | 0.85 (0.24) | 0.12 (0.24) | 0 (0.37) |
| T1-C + T1 + T2 | 0.85 (0.2) | 0.69 (0.38) | 0.06 (0.36) |
| T1-C + T2 + FLAIR | 0.83 (0.31) | 0.79 (0.37) | 0.1 (0.38) |
| T1-C + T1 | 0.82 (0.2) | 0.74 (0.39) | 0 (0.37) |
| T1-C + T2 | 0.85 (0.23) | 0.73 (0.39) | 0.01 (0.38) |
| T1 + T2 | 0.84 (0.28) | 0.2 (0.35) | 0 (0.27) |
| T1 + FLAIR | 0.86 (0.27) | 0.16 (0.35) | 0 (0.3) |
| T1-C + FLAIR | 0.85 (0.2) | 0.09 (0.31) | 0 (0.35) |
| T2 + FLAIR | 0.82 (0.25) | 0.17 (0.32) | 0 (0.29) |

***Table 6.*** *DSC values (median and standard deviation) for different MRI sequence combinations across tumour subregions (T2H, ET, CC). T2H = T2-hyperintense tumour region; ET = enhancing tumour component; CC = cystic component.*

## *4. Discussion*

This study introduces a DL-based framework for the segmentation of PBTs, addressing a critical gap in automated segmentation techniques specifically tailored to paediatric patients. Due to the distinct anatomical and radiological characteristics of paediatric tumours, models trained on adult datasets often fail to generalize effectively. Our approach, leveraging multiparametric MRI sequences (T1, T1-C, T2, and FLAIR), demonstrates human-like performance in whole tumour and tumour subregion segmentation.

Our results are comparable to or slightly higher than previously reported in pediatric segmentation studies(22,23), which achieved DSC ranging from 0.71 to 0.76; and align with a prior paediatric-specific segmentation studies(24). Unlike several existing models that focus solely on specific histologies or tumour locations—thereby limiting their clinical applicability—our model successfully segments a diverse range of tumour histologies and locations, demonstrating adaptability beyond single tumour types types including both intra- and extra-axial tumours, which introduces greater heterogeneity and makes the segmentation task inherently more challenging. Therefore, while our performance is comparable to these studies, it was achieved under substantially more heterogeneous and clinically representative conditions. A fundamental challenge in paediatric neuro-oncology remains the scarcity of large, annotated imaging datasets, which hinders the development of robust and generalizable deep learning models(23). Despite this, our model effectively segments a wide spectrum of tumour subtypes and anatomical locations, showcasing its potential for broader clinical use.

A key contribution of our work is the development of a new annotation protocol specifically tailored to the clinical utility of paediatric neuro-oncology. Unlike the BraTS framework, which includes enhancing tumour, necrosis and oedema—features less relevant for PBTs—we prioritized subregions that are critical for clinical decision-making in children, such as T2H. T2H segmentation demonstrated nearly identical performance to WT (mean DSC 0.85), validating it as a reliable surrogate for volumetric tumour assessment (Figure 2A). This shift in labelling emphasis not only improves segmentation accuracy but also enhances clinical relevance, particularly in longitudinal monitoring scenarios where contrast enhancement may fluctuate or be absent altogether. As such, the adoption of T2H-based segmentation represents a necessary evolution in the development of automated tools for paediatric neuro-oncology.

Despite the robust training framework, segmentation in paediatric neuro-oncology presents unique challenges due to anatomical variability and radiological heterogeneity. To mitigate model sensitivity to false positives—particularly in ET and CC—we implemented a voxel thresholding step during postprocessing. We excluded segmentations smaller than 125 voxels (0.125 mL), which are typically non-clinically relevant and often represent noise. This small-volume threshold, in combination with 26-connectivity filtering, ensures that only spatially coherent and volumetrically significant tumour regions are retained (32,33). Our segmentation framework enables users to adapt this thresholding step based on specific application needs, providing flexibility for clinical versus research use cases.

Another novel aspect of this study is our investigation into how segmentation performance varies by tumour type and patient age. Our analysis revealed that segmentation accuracy differed significantly across tumour histologies and anatomical locations, with ependymomas showing the highest accuracy and LGGs the lowest (Figure 2B). This likely reflects the anatomical variability, less pronounced enhancement, and heterogeneous morphology typical of LGGs. Age-related variability also suggests that developmental anatomy influences model generalizability, underscoring the need for large, diverse training sets. Additionally, similar performance variations were observed across anatomical locations, reflecting structural and imaging complexities unique to each brain region (Figure 2C). This variability highlights the heterogeneous nature of PBTs and the complexity of their radiological features, which can make consistent segmentation more challenging(2,29). It also underscores the need for models capable of handling a wide range of tumour types and brain locations, which this study addresses by training the model on a diverse cohort of patients with different tumour histologies.

In terms of specific tumour subregions, the segmentation of T2H showed excellent performance (Figure 2). This results particularly relevant given that only 63% of patients exhibited enhancing tumour components, and merely 31% presented with a cystic component. The use of T2H ensures a more comprehensive assessment of tumour burden and facilitates better clinical decision-making in the paediatric population.

ET segmentation performed reasonably well, with a mean DSC of 0.75, although some variability was observed due to the differing levels of enhancement across tumours, and also seen in the inter-variability measurement. In contrast, CC segmentation showed lower performance, with a mean DSC of 0. This poor performance can be attributed to the heterogeneous nature of cystic components, which may vary significantly in appearance across different tumour types and patient cohorts(2).

These challenges are consistent with findings from Familiar et al., who highlight the inherent difficulties in distinguishing mild enhancement from non-enhancing tissue and delineating cystic components. Paediatric tumours often exhibit mixed solid-cystic characteristics and bright yet mild enhancement, which complicates segmentation and can introduce inter- and intra-observer variability. As illustrated in Figure 3, variability is minimal in clearly enhancing tumours but increases substantially in mildly enhancing cases. Similarly, the cystic components exhibit diverse imaging appearances, complicating consistent segmentation. We also observed a "Christmas tree effect" in manual segmentations—discontinuous voxels visible in coronal and sagittal planes due to axial-only annotations—underscoring the limitations of manual approaches and the need for automated methods ensuring 3D coherence. This artefact—where isolated voxels appear scattered or stacked across slices, forming a tree-like structure—is likely due to the fact that manual segmentations are typically performed in the axial plane. These variations contribute to the difficulty in training the model, leading to segmentation failures. When evaluating cases with lower DSCs in ET segmentation, two key trends emerged: (1) the model occasionally predicted a few voxels of enhancing tumour when the ground truth contained none, resulting in a DSC of zero, and (2) mildly enhancing regions in the ground truth were sometimes missed by the model. These findings underscore the need for improved classification techniques and more precise ground truth labelling. Familiar et al. also stress the need for standardized imaging protocols and annotation criteria. Variability in imaging acquisition across institutions limits model generalizability, reinforcing the importance of multi-institutional collaboration and standardized data curation.

Our inter- and intra-rater analysis supports the model's reliability, with T2H and whole tumour segmentation showing concordance with human performance(37). The models' lower segmentation performance for ET and CC might be explained by the higher ambiguity in segmenting these subregions, which is also present in human segmentations as reflected by the lower inter- and intra-rater scores for these channels. These findings highlight the potential of AI to reduce variability and improve reproducibility—crucial for clinical adoption.

Manual segmentation of these tumour subregions remains the gold standard; however, this approach is time-consuming, labour-intensive, and subject to inter-operator variability.(7,38) DL-based automation offers a scalable solution, enabling improved tumour tracking, treatment response assessment, surgical planning, and potentially radiomics. Despite promising developments, the clinical use of automated paediatric segmentation has been constrained by limited training datasets(23).

One of the most significant advantages of automated segmentation is the potential reduction in manual contouring time. Our study quantified these time savings, demonstrating an 83% reduction in segmentation time for T2H, 42% for ET, and 27% for CC. These efficiencies are particularly relevant in clinical practice, where radiologists must balance segmentation accuracy with workflow constraints. By minimizing manual intervention while maintaining segmentation quality, deep learning models can enhance productivity and facilitate real-time clinical decision-making. This is especially beneficial for longitudinal tumour monitoring and treatment response assessment, where volumetric changes over time need to be accurately quantified.

We also systematically explored the contribution of different MRI sequences to segmentation accuracy, evaluating multiple sequence combinations to identify clinically efficient and diagnostically effective inputs. Our results highlight that combining T1, T1-C and T2 sequences offers an optimal balance of segmentation performance. This sequence triad delivered high DSC across all subregions, offering guidance for protocol optimization in routine clinical settings, especially when scan time is a concern. Notably, we found that for volume assessment and delineation of tumour boundaries, the inclusion of T1-C did not significantly impact performance—the segmentation accuracy of T2H and WT was essentially the same with or without T1-C. This underscores the limited value of contrast-enhanced sequences for T2-based volumetric assessment, which could further streamline imaging protocols and reduce contrast agent exposure in paediatric patients.

In conclusion, our DL-based segmentation framework represents a significant advancement in paediatric brain tumour segmentation. Importantly, the trained models and pre-processing pipelines have been publicly released to ensure reproducibility and foster future research. In addition, the sharing of anonymized imaging data is under consideration, subject to institutional and ethical approvals, to further accelerate research in this field. By integrating multiparametric MRI and leveraging an advanced 3D nnU-Net architecture, our approach offers a clinically viable tool for volumetric assessment, treatment planning, and response monitoring.

Despite its robust performance, several limitations must be considered. Although trained on a diverse paediatric cohort, the dataset size remains a constraint, potentially affecting the model's generalizability to rarer tumour subtypes or atypical cases. Being a single-center study, sample sizes for some histologies were particularly small, limiting the strength of conclusions for rare subtypes (e.g. meningiomas, craniopharyngiomas). Performance was also lower in low-grade gliomas (LGG), reflecting their high histological and imaging variability, which reduces the generalizability of a universal model across tumour types. Additionally, segmentation of cystic components and enhancing tumours remains challenging, emphasizing the need for improved ground truth annotations and more refined learning strategies, particularly for tumours with subtle enhancement or heterogeneous cystic structures. Another key challenge is the variability in imaging protocols across institutions, and the absence of external validation represents a further limitation, highlighting the importance of future multi-institutional studies. While our model significantly reduces manual segmentation time, expert review is still required for quality control, meaning further validation in real-world clinical settings is necessary before widespread implementation.

To address the comparison with current publicly available models, we additionally benchmarked our network against the first-ranked models of the BraTS Challenge 2023 and 2024. When applied to our dataset, the BraTS-winning model achieved a Dice score of 0.11 for the WT, confirming that current public models fail to perform on skull-included data and across the full range of tumour types. We used WT as our primary evaluation label because our annotation protocol differs from BraTS. As a cross-validation, we ran the same BraTS-winning code on the BraTS training data, reproducing the expected WT performance (0.79). Furthermore, when we tested our network on the BraTS dataset, it achieved a WT Dice of 0.69, notably more adaptable despite never being trained on skull-stripped images and being tested just on HGG cases. This supports that our model generalizes better across anatomical and histological variations than current state-of-the-art BraTS networks.

Future research should focus on expanding training datasets, optimizing model architectures for more challenging tumour subregions, and conducting extensive clinical validation to support broader adoption in paediatric neuro-oncology. For clinical translation, several practical steps are still required. Integration into hospital PACS and radiology reporting workflows is essential to enable seamless use by clinicians, while maintaining strict de-identification and compliance with European data protection regulations. Moreover, formal certification as a medical device (e.g., CE or FDA) and prospective, multi-centre validation would be necessary before deployment in clinical routine. At the current stage, the model is intended for expert-assisted pre-contouring rather than autonomous use. Nonetheless, from a research and software perspective, the framework is already well prepared for translation: the entire pre-processing and inference pipeline is containerized, publicly released on GitHub, and can be integrated into existing research environments or prototype clinical systems.

## *Required statements*

### Ethics

This retrospective study was conducted with approval from the institutional review board of the University Children's Hospital Zurich. Informed consent was obtained from all subjects or their legal guardians at the time of initial diagnosis and imaging, in accordance with the principles outlined in the World Medical Association Declaration of Helsinki.

### Funding

Swiss National Science Foundation (A.P.); Forschungszentrum für das Kind Universität Zürich (A.P.).

### Conflict of Interest

None declared.

### Authorship

- Conceptualization and study design: Arianna Piffer, Ana Guerreiro Stücklin, Bjoern Menze, Florian Kofler, Andras Jakab
- Data curation and annotation: Arianna Piffer, Antonio Giulio Gennari, Patrice Grehten, Selma Sirin, Elizabeth Ross.
- Model development and evaluation: Arianna Piffer, Josef Alois Buchner, Ivan Ezhov, Marcel Rosier, Marie Piraud, Jan Caspar Peeken, Florian Kofler
- Statistical analysis: Arianna Piffer, Josef Alois Buchner, Florian Kofler, Andras Jakab
- Supervision: Ana Guerreiro Stücklin, Bjoern Menze, Florian Kofler, Andras Jakab
- Manuscript writing – original draft: Arianna Piffer
- Manuscript review and editing: All authors (Arianna Piffer, Josef Alois Buchner, Antonio Giulio Gennari, Patrice Grehten, Selma Sirin, Elizabeth Ross, Ivan Ezhov, Marcel Rosier, Jan Caspar Peeken, Marie Piraud, Bjoern Menze, Ana Guerreiro Stücklin, Andras Jakab, Florian)
- Shared last authorship: Florian Kofler, Andras Jakab

### Data Availability

The segmentation models and pre-processing pipeline are publicly available via the BrainLesion GitHub repository (https://github.com/BrainLesion). Anonymized imaging data and annotations used in this study may be made available from the corresponding author upon reasonable request for non-commercial academic purposes.